\documentclass[letterpaper]{article} 
\usepackage{aaai2026}  
\usepackage{times}  
\usepackage{helvet}  
\usepackage{courier}  
\usepackage[hyphens]{url}  
\usepackage{graphicx} 
\usepackage{natbib}  
\usepackage{caption} 
\usepackage{algorithm}
\usepackage{algorithmic}
\usepackage{amsmath}
\usepackage{booktabs}
\usepackage{amsfonts}
\usepackage{multirow}

\usepackage{newfloat}
\usepackage{listings}
\DeclareCaptionStyle{ruled}{labelfont=normalfont,labelsep=colon,strut=off} 
\floatstyle{ruled}
\newfloat{listing}{tb}{lst}{}
\floatname{listing}{Listing}
\title{AutoPP: Towards Automated Product Poster Generation and Optimization}

\author{
    Jiahao Fan\equalcontrib,
    Yuxin Qin\equalcontrib,
    Wei Feng\thanks{Project Leader and Corresponding Author},
    Yanyin Chen,
    Yaoyu Li,
    Ao Ma,
    Yixiu Li,
    Li Zhuang,\\
    Haoyi Bian,
    Zheng Zhang,
    Jingjing Lv,
    Junjie Shen,
    Ching Law
}
\affiliations{
    \textsuperscript{}JD.COM\\

    \{fanjiahao5,qinyuxin11,fengwei25,chenyanyin6,liyaoyu1,maao.8,liyixiu5,zhuangli10,bianhaoyi,zhangzheng11,\\lvjingjing1,shenjunjie,lawching\}@jd.com
}

\usepackage{enumitem}

\begin{document}

\maketitle

\begin{abstract}
Product posters blend striking visuals with informative text to highlight the product and capture customer attention. However, crafting appealing posters and manually optimizing them based on online performance is laborious and resource-consuming. To address this, we introduce AutoPP, an automated pipeline for product poster generation and optimization that eliminates the need for human intervention. Specifically, the generator, relying solely on basic product information, first uses a unified design module to integrate the three key elements of a poster (background, text, and layout) into a cohesive output. Then, an element rendering module encodes these elements into condition tokens, efficiently and controllably generating the product poster. Based on the generated poster, the optimizer enhances its Click-Through Rate (CTR) by leveraging online feedback. It systematically replaces elements to gather fine-grained CTR comparisons and utilizes Isolated Direct Preference Optimization (IDPO) to attribute CTR gains to isolated elements. Our work is supported by AutoPP1M, the largest dataset specifically designed for product poster generation and optimization, which contains one million high-quality posters and feedback collected from over one million users. Experiments demonstrate that AutoPP achieves state-of-the-art results in both offline and online settings. 
\end{abstract}

\begin{links}
\link{Code\&Dataset}{https://github.com/JD-GenX/AutoPP}
\end{links}

\section{Introduction}
Product posters play a crucial role in capturing user interest by artfully combining products, textual content, and visual backgrounds. However, creating posters manually offline and optimizing them through iterative analysis of online performance is both time-consuming and resource-intensive. This challenge has spurred growing interest in automated solutions to streamline both the generation and optimization processes.

\begin{figure*}[t]
    \centering
    \includegraphics[width=0.92\textwidth]{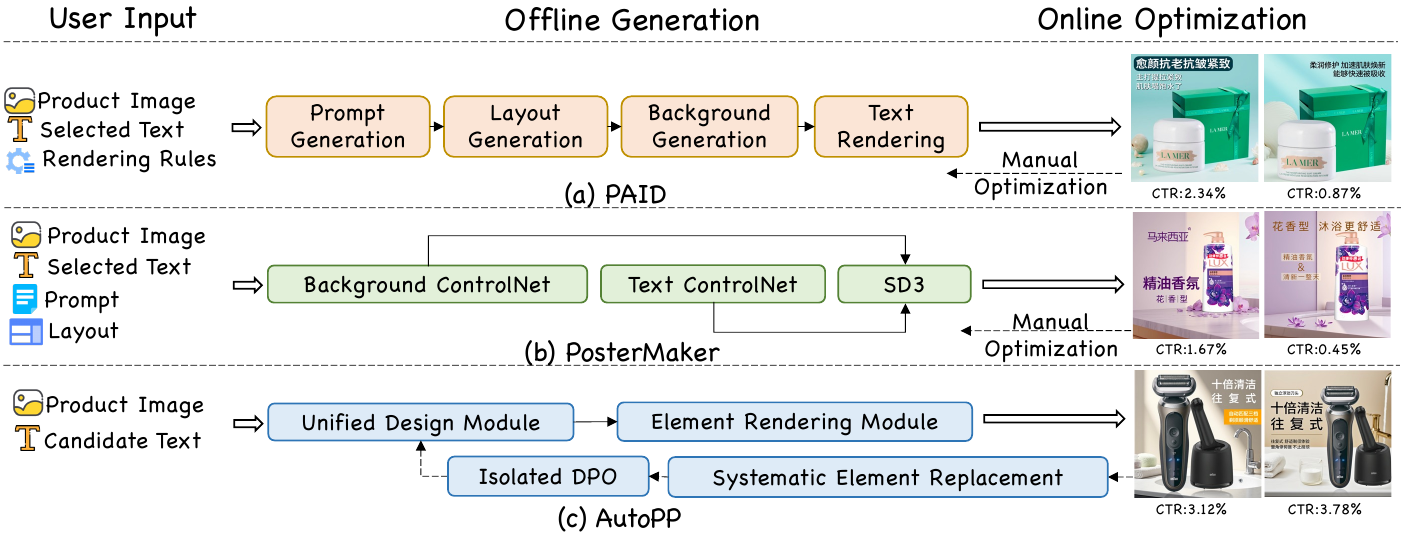}
    \caption{Comparison of different poster generation and optimization pipelines. AutoPP only relies on basic product information and can optimize the generated content automatically based on CTR feedback.}
    \label{fig:pipecmp}
\end{figure*}

With the advancement of text-to-image generation~\cite{rombach2022high,peebles2023scalable,zhang2023adding}, significant progress has been made in automated product poster generation~\cite{gao2025postermaker,wang2024prompt2poster,jia2023cole,chen2025paid}. However, they still require additional manual input at certain stages as shown in Figure~\ref{fig:pipecmp} (a) and (b). PAID~\cite{chen2025paid} employs a four-stage pipeline where three separate models sequentially generate prompts, layouts, and background images. Then, for text rendering, it uses manually defined rules to determine text attributes such as font and color, which inevitably restricts automation and often disrupts the visual harmony between text and background. PosterMaker~\cite{gao2025postermaker} overcomes this issue by utilizing Stable Diffusion 3 (SD3)~\cite{esser2024scaling} with background and text ControlNets~\cite{zhang2023adding}, which automatically renders the text along with background generation. Nevertheless, it requires users to individually craft pleasing layouts and compelling selling points for each poster, making it inefficient for generating high-quality posters at scale. To address these issues, we propose an automated product poster generator as illustrated in Figure~\ref{fig:pipecmp} (c), which requires only basic product information (a product image and candidate text) as input. Our model integrates a unified design module to jointly reason about three key poster elements (background, text, and layout). This comprehensive approach eliminates the fragmentation of task-specific models while ensuring design consistency. We further develop an element rendering module to jointly render the background and text according to the predicted layout. It encodes product images and text into condition tokens, then implements a decomposed attention mechanism with condition self-attention for intra-element modeling and image-condition cross-attention for generation guidance. Our approach reduces the computational overhead along with long token sequences while maintaining robust generation quality.

Despite the automation challenges in offline generation, existing methods also suffer from similar issues in online optimization as shown in Figure~\ref{fig:pipecmp}, often requiring manual adjustment of generation strategies based on empirical observations. Recently, CG4CTR~\cite{yang2024new} and CAIG~\cite{chen2025ctr} have made initial progress by optimizing poster backgrounds for the online metric, like Click-Through-Rate (CTR). However, they represent only a partial solution, since CTR is also heavily influenced by textual content and layout design. A straightforward solution is to extend these methods to optimize background, text, and layout jointly, but this coarse-grained approach cannot indicate which poster element actually drives the CTR improvement, leading to sub-optimal results and low learning efficiency. To address these limitations, we propose a product poster optimizer that automatically refines the content of posters through CTR feedback instead of manual modification. Together with the proposed product poster generator, these two components form \textit{AutoPP}, an automated pipeline that generates and optimizes high-quality product posters for maximum CTR. We first collect CTR feedback on generated sample pairs, each constructed by systematically replacing one element while keeping others constant. Leveraging this fine-grained feedback, we propose Isolated Direct Preference Optimization (IDPO) that attributes CTR improvement to isolated elements. This decomposition enables precise attribution of performance improvements to specific elements, allowing targeted optimization. To support both generation and optimization tasks, we introduce \textit{AutoPP1M}, the largest dataset of product posters, containing one million professionally designed posters and CTR feedback from over one million users. This comprehensive resource enables data-driven model training while providing the necessary performance signals for our optimization framework. To the best of our knowledge, we are the first to unify automated product poster generation and optimization within a single framework.

In summary, our main contributions are as follows:
\begin{itemize}
    \item We propose an automated pipeline \textit{AutoPP} for product poster generation and optimization. It contains a generator that produces high-quality posters with basic product information, eliminating the need for complex manual input.
    \item We introduce an automated optimizer to improve the CTR of generated posters without manual adjustments. With a systematic element replacement strategy and IDPO, we enable element-isolated CTR optimization for posters through controlled pairwise comparisons, overcoming the attribution limitations of holistic approaches.
    \item We present \textit{AutoPP1M}, the largest dataset of product posters, which facilitates new research in automated product poster generation and online performance optimization. Experiments show that our approach achieves state-of-the-art results in both offline and online settings.
\end{itemize}

\section{Related Work}

\subsection{Product Poster Generation}
The task of product poster generation aims to automatically generate visually appealing images that highlight both products and textual content. Existing approaches to product poster generation still struggle to achieve automated production. Recent methods~\cite{lin2023autoposter,jia2023cole,wang2024prompt2poster,jin2022text2poster,liang2024textcengen} rely on complicated processes like layout generation~\cite{lu2025uni,li2023relation}, attribute prediction, slogan creation, and rendering, and require manual tuning or domain-specific rules. While P\&R~\cite{li2023planning} and PAID~\cite{chen2025paid} advance automation by learning to generate layouts directly from product metadata, they critically depend on hand-crafted rules for text rendering.

With the advent of diffusion-based text rendering methods \cite{ma2025glyphdraw2,tuo2024anytext2,liu2024glyph2}, poster text has become more coherent and diverse. TextDiffuser-2~\cite{chen2024textdiffuser} utilizes GlyphControl~\cite{yang2023glyphcontrol} and a language model within the diffusion model to render texts. While these methods have largely unified the poster generation process, they often fail to maintain product appearance consistency. Recently, PosterMaker \cite{gao2025postermaker} has been developed to incorporate product information, featuring background and text ControlNets~\cite{zhang2023adding}. This enables the simultaneous generation of backgrounds and text while preserving product authenticity. However, it still requires users to provide predefined layouts and compelling selling points. Our research aims to achieve fully automated product poster generation, eliminating additional manual input and enhancing efficiency for large-scale creation and rapid deployment.

\subsection{Product Poster Optimization}
As a subtask of image generation, product posters optimization primarily targets offline evaluation metrics~\cite{wang2025generate}, particularly visual aesthetics. ImageReward \cite{xu2023imagereward}, PickScore \cite{kirstain2023pick}, and HPSv2 \cite{wu2023human} leverage human preference data by training reward models on curated text-image pairs, effectively aligning model outputs with human aesthetic judgments. Du et al. \cite{du2024towards} introduce the multimodal Reliable Feedback Network (RFNet) to improve the availability of generated advertising images. Most recently, UinfiedReward-Think \cite{wang2025unified} integrates chain-of-thought reasoning into reward models, enabling the internalization of reasoning processes and proposing the first unified reward model based on multimodal chains of thought.

Since these offline metrics often fail to align with actual user engagement metrics, recent approaches, like CG4CTR~\cite{yang2024new} and CAIG~\cite{chen2025ctr}, optimize for CTR to bridge this gap. CG4CTR develops a CTR-optimized reward model that jointly leverages multimodal features to accurately reflect user click preferences. CAIG advances the reward modeling architecture to a Multimodal Large Language Model (MLLM), enabling a deeper understanding of cross-modal preferences through semantic alignment. However, they only enhance CTR by optimizing poster backgrounds. In contrast, we investigate how multiple elements in poster generation affect CTR and propose IDPO: an optimization framework that decomposes CTR optimization into isolated poster elements.

\begin{figure*}[t] 
    \centering
    \includegraphics[width=1.0\textwidth]{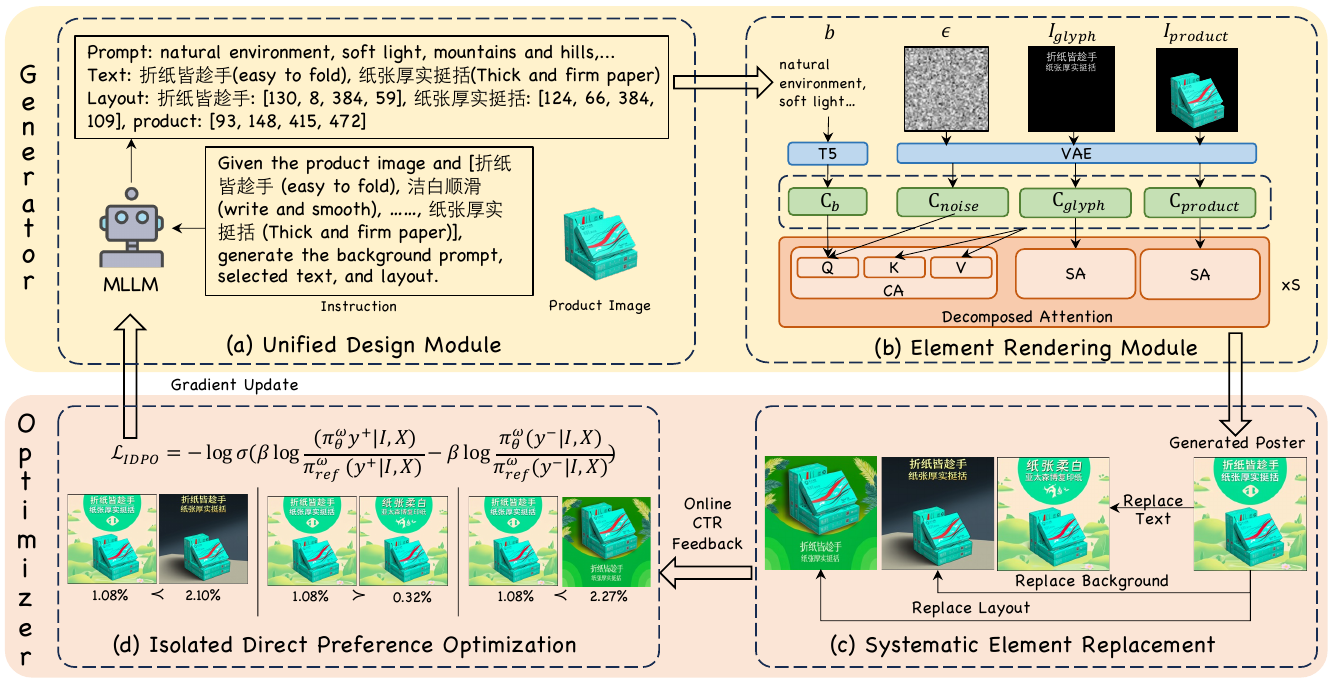}
    \caption{The framework of AutoPP. The generator synthesizes posters while the optimizer refines online CTR. For clarity, we only show the decomposed attention of MM-DiT blocks in the element rendering module. The English translations of the Chinese text are provided in parentheses.}
    \label{fig:method}
\end{figure*}

\section{Dataset}
The AutoPP1M dataset comprises one million product posters, divided into two subsets. The first subset, focused on poster generation, features posters with detailed annotations on visual elements. The second subset, dedicated to poster optimization, includes posters paired with high-quality preference labels.

\subsection{Collection and Annotation}
For the product poster generation subset, we collect product posters from JD.COM, with all posters in a 1:1 aspect ratio and at least 800$\times$800 pixels in size. Starting from an initial candidate pool of approximately 10 million posters, we apply standard data cleaning procedures, including aesthetic filtering, blur detection, and watermark removal \cite{li2024hunyuan, gong2025seedream}, to eliminate low-quality posters, yielding roughly 2 million posters. Subsequently, we employ PaddleOCRv4 \cite{li2022pp} to remove posters containing excessive text or no textual content, resulting in a final refined set of one million posters. This final dataset covers a diverse range of everyday consumer products, spanning over 60 coarse-grained categories. To extract the elements from the poster, we first use Grounding SAM \cite{ren2024grounded} and PaddleOCRv4 to obtain product masks, text content, and text bounding boxes. Then, we generate the background prompt using Qwen2.5-VL \cite{bai2025qwen2}, supplemented with product attribute information (e.g., product names and titles) to mitigate hallucination.

For the product poster optimization subset, we collect CTR feedback for product posters through a 10-day randomized display experiment within JD.COM. To ensure data reliability, each poster is viewed by at least 50 users, accumulating 1,118,140 users. Following the systematic element replacement strategy (detailed in Method), we construct 50,000 pairwise comparisons by matching posters of the same product. Additionally, we ensure a minimum relative CTR difference of 1\% between paired variants to ensure distinguishable CTR differences.

\subsection{Characteristic}
\begin{itemize}
    \item High Quality: The generation subset comprises top-viewed product posters on the platform, reflecting well-designed posters in which merchants invested significant resources. For the optimization subset, the dataset benefits from extensive user engagement on JD.COM, with over one million users, ensuring a high level of diversity and reliability in user preference data.
    \item Large Scale:  Existing product poster datasets, especially those focusing on Chinese product posters, are often limited in scale. With one million elegant product posters, AutoPP1M is an order of magnitude larger than existing datasets \cite{zhou2022composition, lin2023autoposter, hsu2023posterlayout, gao2025postermaker}, enabling more robust training of modern generative models.
    \item New Field: Unlike existing human-feedback image datasets, which typically focus on pure scene images without text \cite{richhf, xu2023imagereward}, AutoPP1M uniquely addresses the significant impact of textual content on user attraction. By providing separate assessments of human preferences for background, text, and layout in posters, we enable more nuanced, element-level modeling of these preferences. This facilitates new research directions in fine-grained preference learning.
\end{itemize}

\section{Method}

\subsection{Overview}
In this work, we propose AutoPP, an automated pipeline for product poster generation and optimization. As illustrated in Figure~\ref{fig:method}, given the basic product information, the generator first generates three key poster elements through the unified design module. Then, the element rendering module synthesizes these elements into the product poster. The optimizer further systematically replaces the isolated elements in the generated poster. Finally, we employ isolated direct preference optimization to improve CTR performance.

\subsection{Automated Product Poster Generator}
\subsubsection{Unified Design Module}
Our unified design module generates coherent poster elements (background prompt, text, and layout) through a single integrated process as shown in Figure~\ref{fig:method} (a). Unlike conventional approaches that rely on separate models for each element, often leading to inconsistent designs, we employ a Multimodal Large Language Model (MLLM) to jointly produce all elements. Formally, given an original product image $I_\text{product}$ and a set of candidate text $T$, we formulate a natural language instruction $X_\text{instr}$ using a template function $f_\text{instr}$:
\begin{equation}
    X_\text{instr} = f_\text{instr}(T).
\end{equation}
A concrete instantiation of $f_\text{instr}$ follows this pattern: \textit{Given the product image and [candidate text], generate the background prompt, selected text, and layout.} [candidate text] is a placeholder.

Then, our MLLM models the joint distribution as an autoregressive sequence:
\begin{equation}
    \pi(y | I_\text{product}, X_\text{instr})=\prod_{i} p(y_{i}| I_\text{product}, X_\text{instr},y_{<i}),
\end{equation}
where $y$ denotes the structured unified design output, containing three key elements: background prompt $b$, selected text $T^*$, and layout $l$, and $y_i$ represents the $i$-th token. The module is trained to minimize the cross-entropy loss.

\subsubsection{Element Rendering Module}
Building upon the outputs from the unified design module, our element rendering module generates the final poster through an efficient token-based architecture as shown in Figure~\ref{fig:method} (b). We first prepare two specialized control signals: a product image $I_{\text{product}}$ positioned on a black background following the predicted product layout, and a glyph image $I_{\text{glyph}}$ with text rendered at the predicted text layout. These mask-form control images provide explicit spatial guidance through foreground-background separation. 

Then, we employ a token-based mechanism for spatial robust glyph control. Note that it eliminates the need for pixel-level alignment between glyph and target images, significantly reducing annotation effort. Specifically, given the target image $I_{\text{target}}$, the glyph image $I_{\text{glyph}}$, the product image $I_{\text{product}}$, and the background prompt $b$, we first encode the visual part with VAE~\cite{kingma2013auto}, and the textual part with T5~\cite{ni2021sentence}:
\begin{equation}
\begin{aligned}
    C_x &= \text{VAE}(I_x) \in \mathbb{R}^{N \times D}, & x \in \{\text{target}, \text{glyph}, \text{product}\} \\
    C_b &= \text{T5}(b) \in \mathbb{R}^{M \times D},
\end{aligned}
\end{equation}
where $C_x$ and $C_b$ represent the encoded tokens. $D,N,M$ denote the latent dimension, the flattened sequence length of image patches, and the tokenized text length, respectively. Since $I_{\text{target}}$, $I_{\text{glyph}}$, and $I_{\text{product}}$ share the same resolution, their sequence lengths are equal to $N$. Following the flow matching paradigm, we gradually perturb the target latent $C_{\text{target}}$ through normalized time steps $t \in [0,1]$:
\begin{equation}
C_{\text{noise}} = (1 - t) C_{\text{target}} + t \epsilon,
\end{equation}
where $\epsilon \sim \mathcal{N}(0, \mathbf{I})$ denotes Gaussian noise sampled from a standard normal distribution.

To effectively leverage the encoded token sequence described above, a straightforward approach is to concatenate all tokens and process them jointly through $S$ MM-DiT blocks~\cite{flux2024}. However, this naive strategy leads to quadratic computational costs, making it impractical for long sequences. To alleviate this problem, we modify the full attention in MM-DiT blocks into a decomposed attention (DA) mechanism, which contains two independent components: condition self-attention (SA) and image-condition cross-attention (CA). Specifically, glyph and product tokens are each processed with condition SA, allowing intra-element dependencies to be efficiently captured. To enable information exchange across image tokens and condition tokens, we perform image-condition CA, where the query is constructed by concatenating the prompt and noise tokens, while the key and value are formed by concatenating all types of tokens:

\begin{equation}
\begin{aligned}
    Q &= [Q_b;\; Q_{\text{noise}}], \\
    K &= [K_b;\; K_{\text{noise}};\; K_{\text{glyph}};\; K_{\text{product}}], \\
    V &= [V_b;\; V_{\text{noise}};\; V_{\text{glyph}};\; V_{\text{product}}], \\
\end{aligned}
\end{equation}
where $Q$, $K$, and $V$ denote query, key, and value in the cross-attention mechanism. 

The training objective requires careful design to ensure precise text rendering—a critical requirement in poster generation. In addition to the core flow matching loss, we introduce an OCR perceptual loss that explicitly enforces legibility in text regions:
\begin{equation}
\mathcal{L}_{\text{render}} = \|v_\text{pred} - v_\text{gt}\|_2^2 + \lambda \|f_{\text{ocr}}(I_{\text{pred}}) - f_{\text{ocr}}(I_{\text{target}})\|_2^2,
\end{equation}
where $v_\text{pred}$ and $v_\text{gt}$ are the predicted velocity field and the ground-truth velocity field at each denoising step, respectively. $I_{\text{pred}}$ is the reconstructed image, $f_{\text{ocr}}$ denotes intermediate features from the pretrained PaddleOCRv4 backbone, and $\lambda=0.1$ balances the loss terms.

\subsection{Automated Product Poster Optimizer}

\subsubsection{Systematic Element Replacement}
To assess the contribution of isolated elements to CTR improvement, we develop a strategy that systematically replaces elements in the poster, as shown in Figure~\ref{fig:method} (c). Our approach begins with a generated poster $P$ and its three key elements $y$: background prompt $b$, selected text $T^*$, where $T^* \subseteq T$, and layout $l$. To disentangle the impact of isolated elements, we create modified variants by perturbing individual elements while maintaining others fixed. Specifically, for background replacement, we provide the original product image and prompt $b$ for GPT-4o~\cite{hurst2024gpt}, and require it to generate a distinct prompt $b'$ that is not a repetition of $b$, but is suitable for the product. Text content is replaced with length-matched alternative texts $T^{*'}$ from the same candidate text set $T$. For layout replacement, we regenerate another layout $l'$ by our unified design module. After replacing each element, we use the element rendering module to regenerate a product poster $P'$ with the replaced element $y'$.

\subsubsection{Isolated Direct Preference Optimization}
These posters then undergo a randomized display experiment on JD.COM to collect CTR feedback as shown in Figure~\ref{fig:method} (d). For simplicity, we denote the product image $I_{\text{product}}$ as $I$, the instruction $X_{instr}$ as $X$ in the following description. The experimental result establishes a preference relation $P^+ \succ P^-$, where $P^+$ and $P^-$ denote the higher-CTR and lower-CTR posters between $(P, P')$, respectively. Similar to CAIG~\cite{chen2025ctr}, we focus solely on fine-tuning the unified design module to choose higher attractive elements $y^+$ and reject less attractive elements $y^-$. The standard DPO~\cite{rafailov2023direct} is a coarse-grained alignment method that directly optimizes the overall output of our unified design model $\pi_\theta(y|I,X)$. Given the policy model $\pi_{\mathrm{\theta}}$ and the reference model $\pi_{\mathrm{ref}}$, the DPO objective is formulated as:
\begin{equation}
    \scalebox{0.83}{$\displaystyle
    \mathcal{L_{\text{DPO}}}=-\log\sigma\left(\beta\log\frac{\pi_\theta(y^+|I,X)}{\pi_{\mathrm{ref}}(y^+|I,X)}-\beta\log\frac{\pi_\theta(y^-|I,X)}{\pi_{\mathrm{ref}}(y^-|I,X)}\right),
    $}
\end{equation}
where $\sigma$ denotes the sigmoid function and $\beta$ is a temperature parameter controlling the deviation from the reference model. 


We argue that background $b$, text $T^*$, and layout $l$ contribute differently to user engagement. This motivates our IDPO, which incorporates CTR feedback through element-aware optimization:
\begin{equation}
\begin{split}
     \log\pi^w(y| I,X) &= \frac{\sum_{i} w_i \log p(y_{i}|I,X,y_{<i})}{\sum_{i} w_i},
     \\ w_i &= \sum_{c\in{b,T^*,l}} \alpha_c \cdot \mathbb{I}(y_i \in c),
\end{split}
\end{equation}
where $w_i$ are element-specific weights and $\alpha_c$ controls the importance of element $c$. $\mathbb{I}(\cdot)$ is an indicator function. For instance, emphasizing background modifications would set $\alpha_b > 1 , \alpha_l = \alpha_{T^*}=1$. Additionally, the likelihood is normalized by $\sum_{i} w_i$ to maintain scale consistency. We substitute $\log\pi(y| I,X)$ in the DPO loss with $\log\pi^w(y| I,X)$ to formulate the IDPO loss. This decoupled approach ensures CTR feedback accurately guides modifications to the most influential element, yielding more efficient and interpretable poster optimization compared to holistic approaches.

\begin{figure*}[t] 
    \centering
    \includegraphics[width=0.9\textwidth]{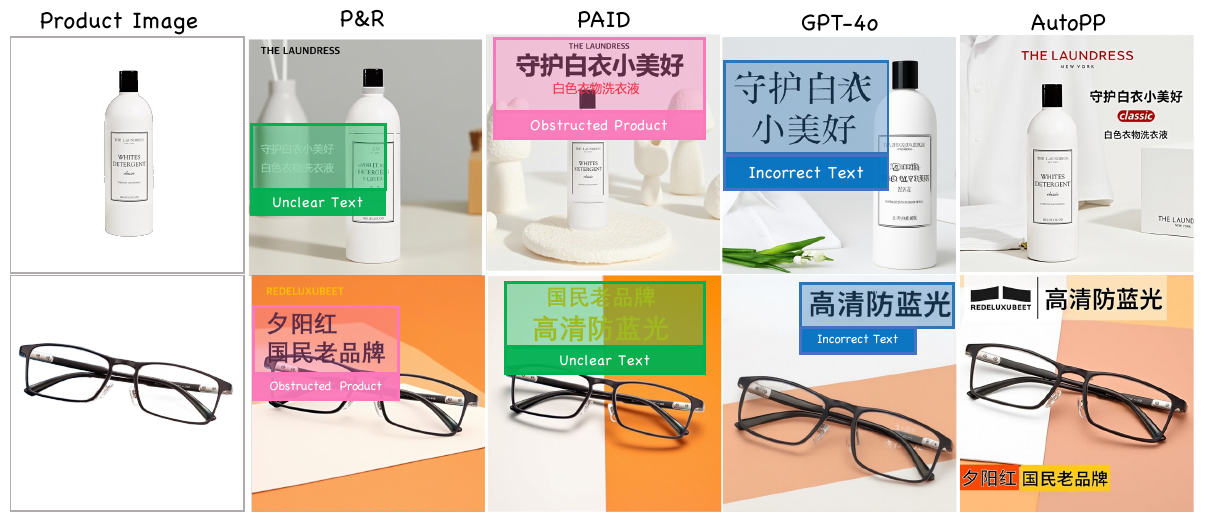}
    \caption{Qualitative comparison with SOTA product poster generation methods. They often suffer from issues such as obstructed products, unclear text, and incorrect text, noted by colorful boxes.}
    \label{fig:sota}
\end{figure*}

\section{Experiment}

\subsection{Implementation Details}
For the product poster generator, we initialize the unified design module from LLaVA~\cite{liu2023visual} and the element rendering module from FLUX.1 dev~\cite{flux2024}. The design module undergoes full-model fine-tuning for 10 epochs using a cosine learning rate scheduler (initialized as 2e-6), while the rendering module employs LoRA~\cite{hu2022lora} adaptation with a learning rate of 5e-6 for 2 epochs, and is trained at the resolution of 800$\times$800. The training phase takes approximately 1 and 2.5 days respectively. The product poster optimizer is trained using our proposed IDPO for 3 epochs, with hyperparameter $\beta=0.5$, $\alpha=1$ for the constant elements, and $\alpha=5$ for the replaced element, taking about 12 hours to complete. All experiments are conducted on a single node with 8 NVIDIA H100 GPUs.

\begin{figure*}[t] 
    \centering
    \includegraphics[width=0.9\textwidth]{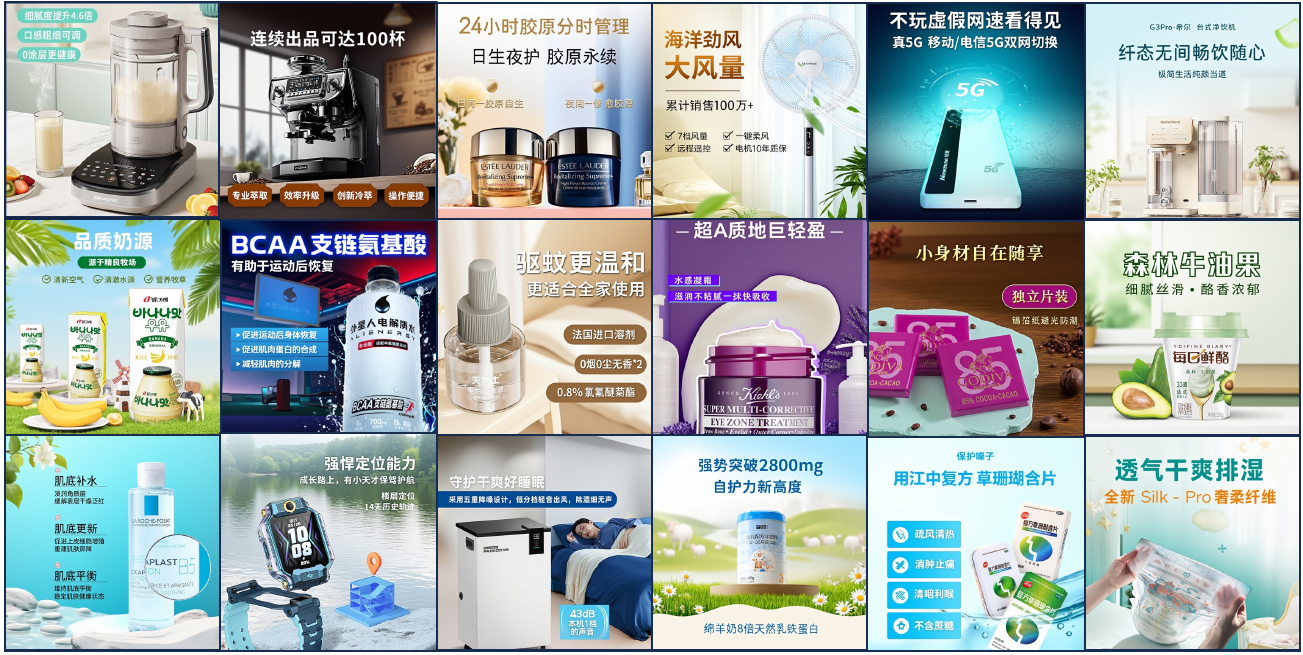}
    \caption{Posters generated by AutoPP. Our method generates diverse layouts for various product categories.}
    \label{fig:case}
\end{figure*}

\begin{table}[h]
    \centering
    \small
    \renewcommand{\arraystretch}{1.15}
    \begin{tabular}{lcccccc}
        \toprule
        \textbf{Method} & \textbf{FID} $\downarrow$ & \textbf{Clip-T} $\uparrow$ & \textbf{Ali} $\downarrow$ & \textbf{Ove} $\downarrow$ & \textbf{MIoU} $\uparrow$  \\
        \midrule
        P\&R            & 104.05 & 27.21 & 0.014 & 0.024 & 0.203  \\
        PAID            & 83.55  & 28.92 & 0.013 & 0.041 & 0.215  \\
        GPT-4o          & 63.47  & 29.58 & 0.009 & 0.018 & 0.140  \\
        \textbf{AutoPP}   & \textbf{60.71} & \textbf{29.75} & \textbf{0.007} & \textbf{0.011} & \textbf{0.256}  \\
        \bottomrule
    \end{tabular}
    \caption{Comparison of product poster generation methods.}
    \label{tab:new_performance}
\end{table}

\begin{table}[h]
    \centering
    \small
    \renewcommand{\arraystretch}{1.15}
    \setlength{\tabcolsep}{4.5pt}
    \begin{tabular}{lcccc}
        \toprule
        \textbf{Method} & \textbf{Sen. Acc} $\uparrow$ & \textbf{NED} $\downarrow$ & \textbf{FID} $\downarrow$ & \textbf{Clip-T} $\uparrow$ \\
        \midrule
        FLUX-Fill    & 33.14 & 49.07 & 51.53 & 28.51 \\
        FLUX-ControlNet & 51.20 & 26.25 & 47.74 & 30.31 \\
        Glyph-byt5-v2       & 52.89 & 14.99 & 54.80 & 30.23 \\
        AnyText-v2       & 54.22 & 21.00 & 73.71 & 29.74 \\
        PosterMaker      & 57.87 & 21.93 & 49.76 & 30.43 \\
        \textbf{AutoPP} & \textbf{65.19} & \textbf{12.94} & \textbf{43.19} & \textbf{30.49} \\
        \bottomrule
    \end{tabular}
    \caption{Comparison of visual text generation methods.}
    \label{tab:performance}
\end{table}

\begin{figure}[ht]
    \includegraphics[width=\linewidth]{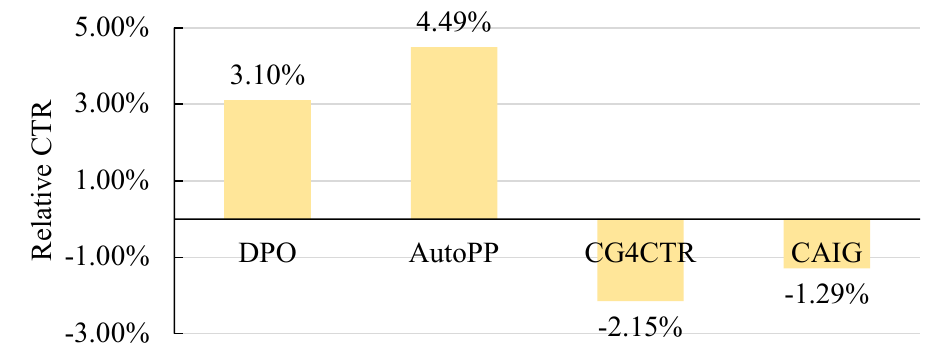}
    \caption{Relative CTR improvement compared with SOTA.}
    \label{fig:pop_cmp}
\end{figure}

\subsection{Product Poster Generation Performance}

\subsubsection{Evaluation Metric}
We evaluate our model from three perspectives on 500 product posters. For overall poster quality, we use FID \cite{heusel2017gans} and CLIP-T \cite{gao2025postermaker} to assess the visual quality of the generated posters. For layout rationality, we adopt Alignment (Ali) to measure how well the elements within the layout align with each other, Overlap (Ove) to calculate the area of overlap between any two elements, and Maximum IoU (MIoU) to evaluate the intersection over union between the generated layout and the ground-truth layout. We assess text rendering using sentence accuracy (Sen. Acc) and normalized edit distance (NED), employing PaddleOCRv4 to detect the text in the generated posters and compare the detected text with the ground-truth text. 

\subsubsection{Comparison with SOTA}
We compare AutoPP with both open-source methods~\cite{li2023planning,chen2025paid} and closed-source GPT-4o~\cite{hurst2024gpt}, where open-source methods are trained on the same dataset as ours. Seedream 2.0~\cite{gong2025seedream} is excluded from the comparison due to its potential to alter product appearances in ways that could mislead consumers. For GPT-4o, we provide it with the same basic product information as ours. As shown in Table~\ref{tab:new_performance}, our method outperforms all others in terms of visual quality and layout rationality. We also showcase several examples in Figure~\ref{fig:sota}. P\&R and PAID may cause unclear text and occlusion of products due to manual rendering rules. GPT-4o performs poorly in Chinese text generation and is prone to stroke errors in text. Some generated posters are shown in Figure~\ref{fig:case}, demonstrating that our method can generate diverse and exquisite product posters.

To further evaluate visual text generation performance, we use ground-truth layouts, background prompts, and text as inputs for all methods. For Glyph-byt5-v2~\cite{liu2024glyph2} and Anytext-v2~\cite{tuo2024anytext2}, inpainting is applied to ensure that the product region remains unchanged. The results are summarized in Table~\ref{tab:performance}. Our method achieves the best performance for both textual accuracy and visual quality metrics, proving the effectiveness of AutoPP in controllable visual text generation.

\subsubsection{Ablation Study} We conduct experiments to validate the efficient and robust generation capability of the element rendering module.

\textit{Efficient Generation:} Thanks to the flexible token mechanism, our method does not add additional parameters when introducing more control conditions. In contrast,  PosterMaker~\cite{gao2025postermaker} and Flux-ControlNet~\cite{zhang2023adding} increase the parameter count by 1.6B and 4.2B, respectively. We further compare DA and full attention's cost. Results show that DA reduces GFLOPs of the MM-DiT block by 18$\%$ (800$\times$800 resolution) and by 24$\%$ (1024$\times$1024 resolution) with comparable performance (65.19 vs 66.37 Sen. Acc↑, 43.19 vs 44.28 FID↓).

\textit{Robust Generation:} Our method directly encodes the glyph image as tokens, making it robust to the misalignment between the glyph image and the product poster.
As shown in the first two rows of Table~\ref{tab:performance}, both Flux-ControlNet and Flux-Fill require pixel-level alignment, resulting in lower Sen. Acc and higher NED compared to our method.

\subsection{Product Poster Optimization Performance}

\subsubsection{Evaluation Metric} We first use the Relative CTR to measure the CTR improvement compared with the non-optimized pretrained product poster generator. Then the Reward Accuracy quantifies the proportion of higher CTR posters that also receive higher model preference probabilities than their lower-CTR counterparts.



\subsubsection{Comparison with SOTA}
To assess the CTR improvement capability of our product poster optimizer, we conduct an evaluation between our AutoPP framework and two state-of-the-art CTR optimization approaches: CG4CTR~\cite{yang2024new} and CAIG~\cite{chen2025ctr}. We randomly select 10,000 products and generate one poster per product using each of the three methods. The CTR performance is collected through a one-week online experiment on JD.COM. As shown in Figure~\ref{fig:pop_cmp}, AutoPP achieves significant improvement in CTR performance, while comparison methods CG4CTR and CAIG show negative Relative CTR due to their failure to incorporate layout and text elements.

\subsubsection{Ablation Study}
We conduct experiments to evaluate the effectiveness and impact of data volume for the optimizer.

\textit{Effective Optimization}: Compared with the pretrained model, AutoPP achieves superior CTR (4.49$\%$) through its automated online optimization capability, as shown in Figure~\ref{fig:pop_cmp}. More importantly, while both AutoPP and DPO utilize the same training data, AutoPP’s isolated optimization strategy enables more effective learning of the complex relationships between poster elements, leading to better CTR performance (4.49$\%$ vs. 3.10$\%$). Notably, even mere 0.5$\%$ CTR enhancements can yield significant financial returns, underscoring the practical value of our method.

\textit{Data Volume Impact}: We first train the product poster optimizer across the range 10K, 30K, 50K samples and calculate Reward Accuracy on 1,000 test samples. Reward accuracy shows consistent improvement with increasing data volume (51.20$\%$ → 67.19$\%$ → 75.99$\%$), confirming the framework’s ability to effectively leverage larger datasets without performance saturation.

\section{Conclusion}
In this work, we introduce AutoPP, an automated pipeline for high-quality product poster generation and optimization. It integrates background, text, and layout design into a unified module, followed by an element rendering module, enabling generation with basic product information while overcoming the limitations of fragmented workflows. To further enhance online performance, we propose a systematic element replacement strategy coupled with IDPO, specifically designed for maximizing CTR. To advance research in the related field, we will release AutoPP1M—the largest dataset for poster generation and optimization to date. Experiments validate that AutoPP achieves SOTA results in both offline and online settings. 

\bibliography{aaai2026}

\newpage

\section{Appendix}

This supplementary material provides:
\begin{enumerate}[label=(\arabic*), leftmargin=*, align=left]
    \item \textbf{Section A.1.} Additional results generated by AutoPP under diverse configurations.
    \item \textbf{Section A.2.} Comparison of parameters with ControlNet-based poster generation models.
    \item \textbf{Section A.3.} Representative examples of the instructions used for GPT-4o.
    \item \textbf{Section A.4.} Discussion on current limitations of the method and proposed directions for future research.
    \item \textbf{Section A.5.} Analysis of potential social impacts, ethical considerations, and safeguards implemented.
\end{enumerate}

\subsection{A.1 More Visualization Results}
\subsubsection*{A.1.1 Comparison with Visual Text Generation Models}

To establish a rigorous comparison framework, we maintain strict control over experimental variables by keeping the product, layout, and text constant across all trials. This controlled setup enables focused evaluation of text rendering performance across different methods. As demonstrated in Figure~\ref{fig:compare-text}, AutoPP exhibits superior performance in text generation quality compared to existing approaches, particularly in handling challenging small-font rendering scenarios. 

\subsubsection*{A.1.2 Comparison with FLUX variants}

\begin{figure*}[t] 
    \centering
    \includegraphics[width=\textwidth]{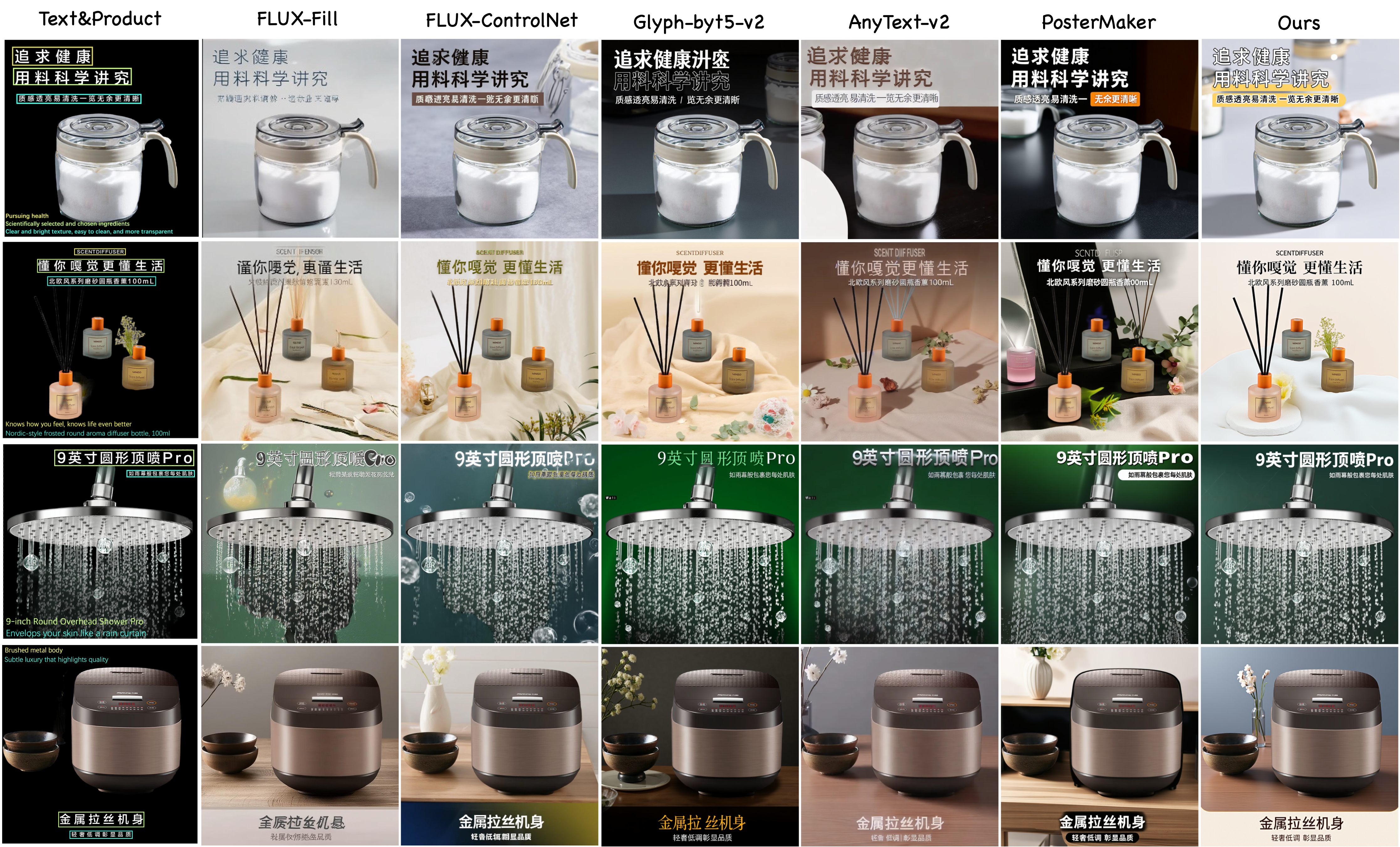}
    \caption{Comparison with SOTA text rendering methods. We combine the product image and glyph image in the first column, and additionally translate the Chinese text into English, rendering it above in different colors. Our method outperforms existing models in text generation accuracy.}
    \label{fig:compare-text}
\end{figure*}

We evaluate two FLUX variant implementations under the same controlled conditions outlined in Section~A.1.1. The visualization results (columns 2-3 in Figure~\ref{fig:compare-text}) reveal fundamental limitations in both FLUX-ControlNet and FLUX-Fill architectures. Due to the loose spatial alignment between the glyph image and the target image, both methods encounter significant optimization challenges, which manifest visually as compromised text legibility.

\subsubsection*{A.1.3 More Products}

\begin{figure*}[t] 
    \centering
    \includegraphics[width=\textwidth]{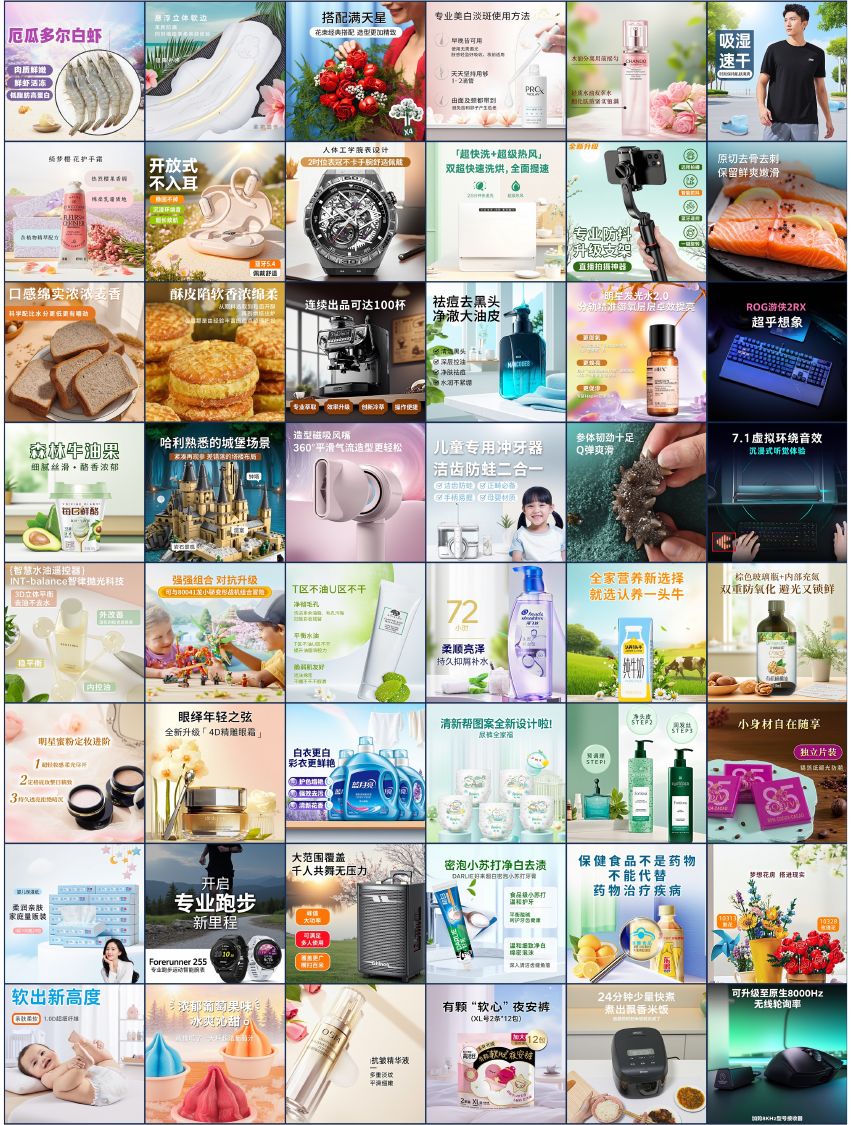}
    \caption{More posters generated by AutoPP. Our method is capable of generating diverse layouts for a wide range of product categories.}
    \label{fig:case-appendix}
\end{figure*}

In Figure~\ref{fig:case-appendix}, we present additional poster generation results produced by AutoPP across a broader range of products. AutoPP consistently produces visually appealing posters for diverse product categories. Notably, the text color in the posters is well harmonized with the overall theme, and in some cases, the model generates an underlay for the text, further enhancing its visual prominence.

\subsubsection*{A.1.4 More Aspect Ratios} 

\begin{figure*}[t] 
    \centering
    \includegraphics[width=\textwidth]{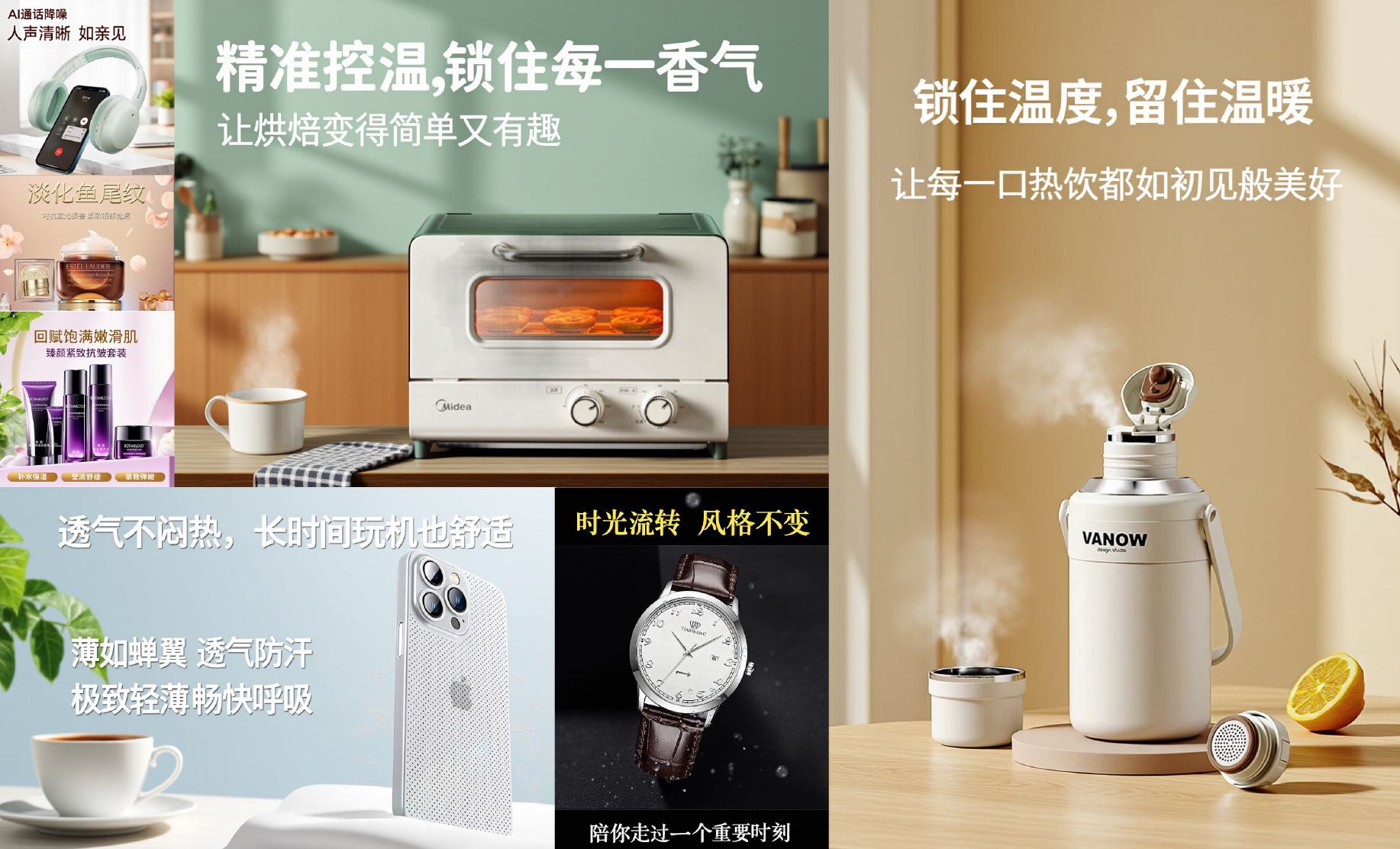}
    \caption{Examples with more aspect ratios. Our model supports poster generation in multiple aspect ratios, such as 4:3, 16:9.}
    \label{fig:multi-aspect}
\end{figure*}

Benefiting from FLUX's ability to generate images at various resolutions, AutoPP inherits this capability and maintains high-quality generation across different aspect ratios. As evidenced in Figure~\ref{fig:multi-aspect}, AutoPP maintains typographic fidelity and layout coherence across five challenging aspect ratios, addressing critical industry requirements for responsive visual design. 

\subsubsection*{A.1.5 More Languages}
\begin{figure*}[t] 
    \centering
    \includegraphics[width=\textwidth]{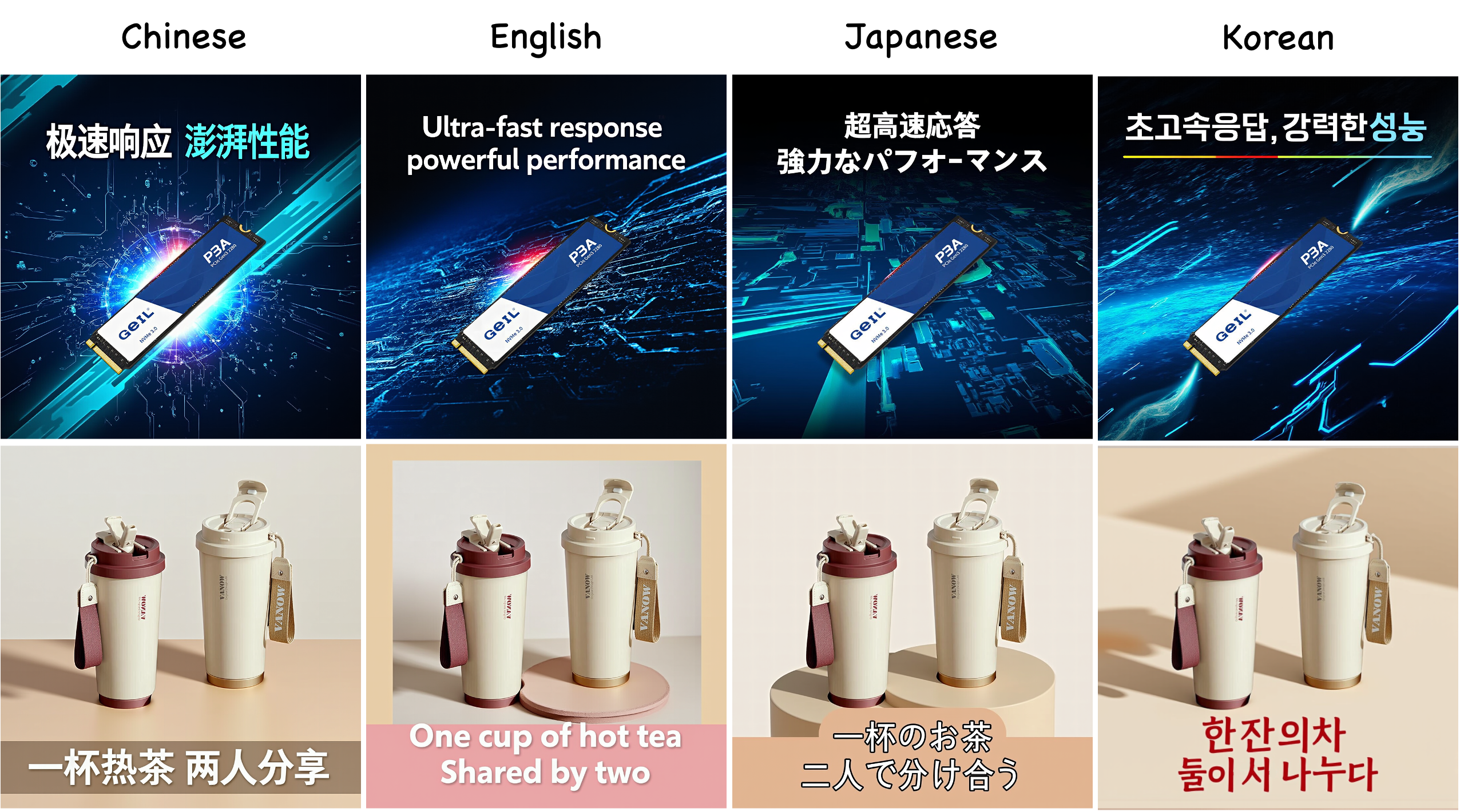}
    \caption{Examples with more languages. We generate glyph images using Chinese, English, Japanese, and Korean texts, and then create posters based on these multilingual glyph images.}
    \label{fig:multi-language}
\end{figure*}

Despite being mainly trained on Chinese characters, AutoPP shows emergent cross-lingual generalization. As shown in Figure~\ref{fig:multi-language}, it can generate valid glyphs for English, Japanese, and Korean characters not seen during training.

\subsubsection*{A.1.6 More Layouts}
\begin{figure*}[t] 
    \centering
    \includegraphics[width=\textwidth]{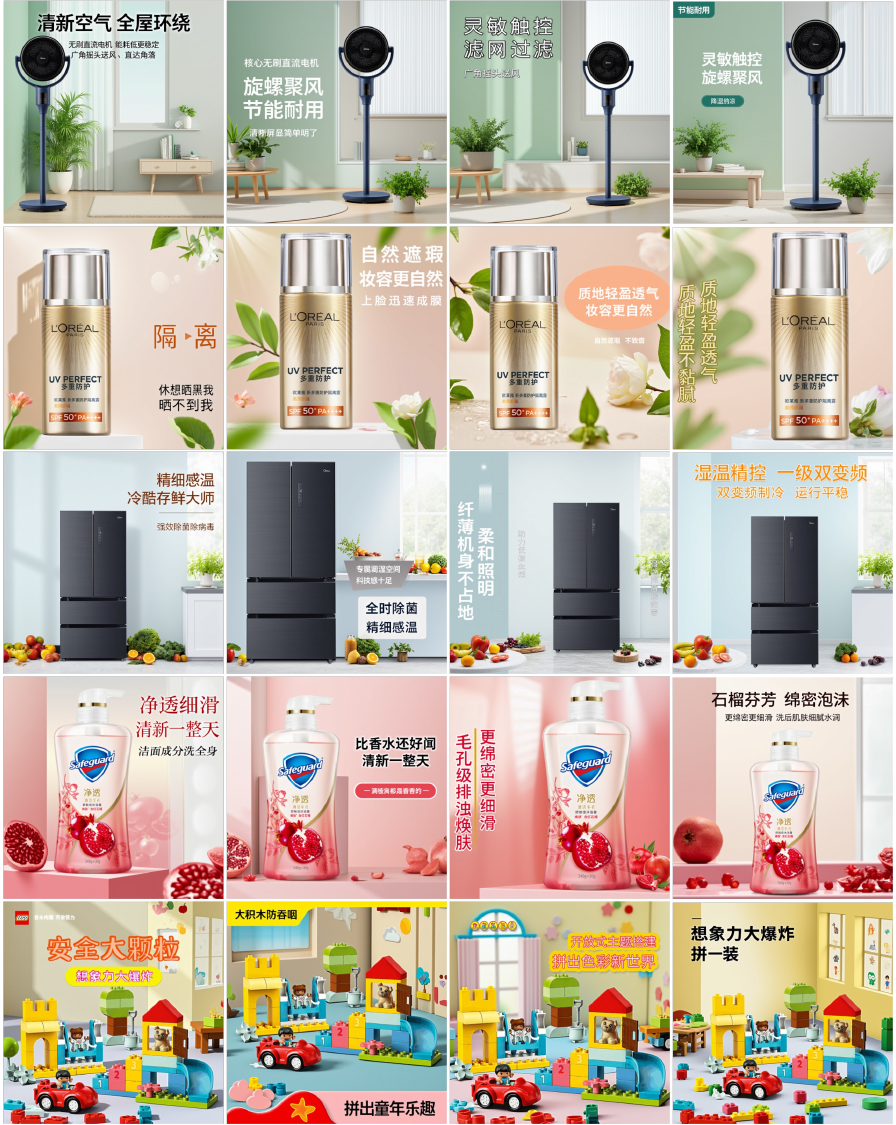}
    \caption{Multiple layout results. Each row shows poster results for the same product with different layouts.}
    \label{fig:multi-layout}
\end{figure*}

For the same product, AutoPP can generate posters with multiple layouts. Given a product and candidate texts, we utilize the unified design module to produce diverse layouts, and then render posters based on these different arrangements. As illustrated in Figure~\ref{fig:multi-layout}, AutoPP generates four distinct layouts and corresponding texts for each product. This multi-layout generation mechanism can satisfy the preferences of different users.

\subsubsection*{A.1.7 Training Samples}
\begin{figure*}[t] 
    \centering
    \includegraphics[width=\textwidth]{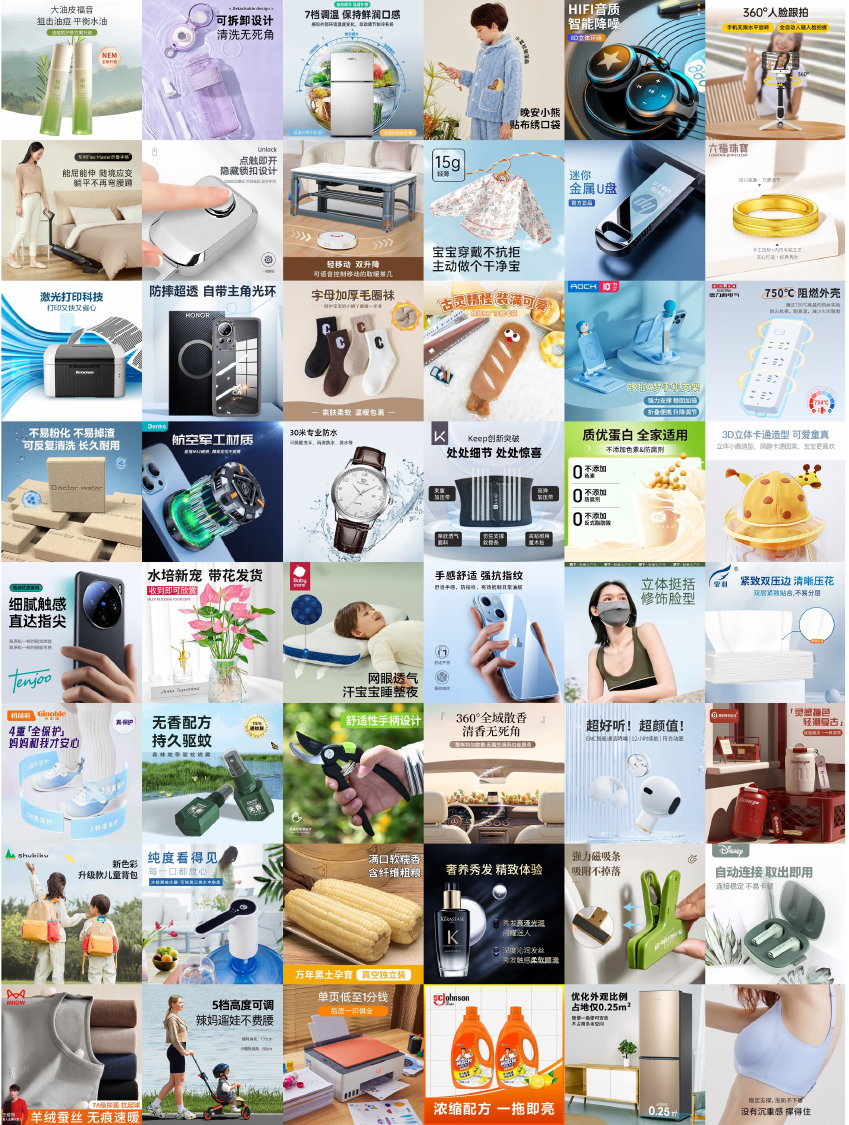}
    \caption{Samples from the AutoPP1M generation subset.}
    \label{fig:dataset}
\end{figure*}

Figure~\ref{fig:dataset} showcases example training images from our dataset, which primarily features Chinese text, with a minor inclusion of English text. The dataset also contains challenging scenarios involving small-sized text elements.

\subsection{A.2 Comparison of Model Parameters}
We compare AutoPP with PosterMaker and Flux-ControlNet. Unlike these methods, AutoPP adds no extra parameters for product and text inputs, while PosterMaker and Flux-ControlNet add 1.6 billion and 4.2 billion parameters, respectively, as shown in Table~\ref{tab:comparison}.

\subsection{A.3 Instruction Set}
In our experiments, we employ the GPT-4o model to perform poster generation. We provide it with product images and candidate text, and then use carefully designed instructions to guide the generation process. A selection of these instructions is presented in Table~\ref{tab:7rows_merged}.

\subsection{A.4 Limitations and Future Work}
This study optimizes CTR using aggregated data from all users, which may overlook the preferences of minority groups and lead to suboptimal user experiences. In future work, we plan to investigate personalized human feedback learning mechanisms to more finely explore how individual preferences affect product poster generation. We also aim to integrate the design and rendering module into a single autoregressive model, enabling unified optimization with RLHF. Through these efforts, we seek to develop a more flexible architecture for product poster generation.

\subsection{A.5 Social Impact}
Our research on automated product poster generation offers several positive impacts. Firstly, it enhances enterprise efficiency through a fully automated workflow. Secondly, it supports small businesses by reducing technical barriers and promoting fair competition in marketing. We ensure ethical use by clearly labeling AI-generated images and adhering to commercial guidelines. 

\begin{table}[h]
    \centering
    \small
    \renewcommand{\arraystretch}{1.15}
    \setlength{\tabcolsep}{4.5pt}
    \begin{tabular}{lccc}
        \toprule
        Method & Backbone & Product branch & Text branch\\
        \midrule
        PosterMaker & 2.3B & 0.8B & 0.8B \\
        Flux-ControlNet & 12B & 2.1B  & 2.1B\\
        Ours & 12B & - & -\\
        \bottomrule
    \end{tabular}
    \caption{Comparison of model parameter counts.}
    \label{tab:comparison}
\end{table}


\begin{table*}[h!]
    \small
    \renewcommand{\arraystretch}{1.5}
    \begin{tabular}{|l|p{16cm}|} 
        \hline
        \multirow{7}{*}{Instructions} & \textit{Using the provided product image, create a 1:1 product poster that must include the text {}. The design should feature vibrant color contrasts, dynamic composition, and a modern aesthetic to maximize visual appeal, aligning with cutting-edge marketing trends to drive customer engagement.} \\
        & \textit{Design a square (1:1) product advertisement poster based on the given product image. Incorporate the text: {} into the poster. The final image should be attractive and meet professional marketing design standards.} \\
        & \textit{Create a premium 1:1 product poster using the provided product image as the main visual. Design the poster to match the product’s brand tone, and ensure it includes the following text: {}. The design should highlight the product’s quality.} \\
        & \textit{Based on the given product photo, generate a 1:1 poster incorporating the text {}. Focus on minimalist design—clean layouts, subtle color palettes, and ample white space—while ensuring the product and text stand out harmoniously. The result should exude sophistication, fitting high-end product marketing strategies.} \\
        & \textit{Using the supplied product photo, generate a 1:1 ratio promotional poster. The poster should prominently display the text: {}. Make sure the layout is engaging and effective for marketing purposes.} \\
        & \textit{Generate a marketing-style 1:1 product poster with the product image centered. Display the text {} in a suitable and prominent position on the poster, ensuring it is clearly visible without overpowering the product image. Use simple and elegant typography with a limited color palette to emphasize a modern marketing atmosphere. } \\
        & \textit{Using the provided product visuals, produce a 1:1 poster that includes the text {}. Emphasize bold typography for the text, paired with eye-catching graphic elements (e.g., gradients, geometric shapes) to create a striking focal point. The overall look should be energetic and attention-grabbing, suitable for promotional campaigns aiming to boost brand visibility.} \\
        \hline
    \end{tabular}
    \caption{Instruct directives for GPT-4o.}
    \label{tab:7rows_merged}
\renewcommand{\arraystretch}{1}
\end{table*}

\end{document}